\documentclass{article}
\usepackage{times}
\usepackage{graphicx}
\usepackage{subfigure}
\usepackage{natbib}
\usepackage{algorithm}
\usepackage{algorithmic}
\usepackage{hyperref}
\usepackage{times}
\usepackage{epsfig}
\usepackage{stmaryrd}
\usepackage[mathscr]{eucal}
\usepackage{lineno}
\usepackage{color}
\usepackage{filecontents}
\usepackage{subfigure}
\usepackage{multirow}
\usepackage{verbatim} 
\usepackage{tabularx}
\usepackage{multirow}
\usepackage{textcomp,booktabs}
\usepackage{lineno}
\usepackage{stmaryrd}
\usepackage[mathscr]{eucal}
\usepackage{amsmath}
\usepackage{amssymb}

\def\C{{\bf C}}

\def\D{{\bf D}}

\def\h{{\bf h}}

\def\H{{\bf H}}
\def\G{{\bf G}}

\def\I{{\bf I}}

\def\X{{\bf X}}

\def\Q{{\bf Q}}

\def\x{{\bf x}}

\def\z{{\bf z}}

\def\m{{\bf m}}
\def\n{{\bf n}}

\def\W{{\bf W}}

\def\0{{\bf 0}}
\def\1{{\bf 1}}

\def\RB{{\mathbb R}}

\def\Si{\mbox{\boldmath$\Sigma$\unboldmath}}

\def\eg{\emph{e.g. }}
\def\ie{\emph{i.e. }}

\def\argmin{\mathop{\rm argmin}}

\def\etal{{\em et al.\/}\,}

\graphicspath{{../figureCORR/}}   

\usepackage[accepted]{icml2014}

\icmltitlerunning{ Learning Mid-Level Features and Modeling Neuron Selectivity for Image Classification}

\begin{document}

\twocolumn[
\icmltitle{ Learning Mid-Level Features and Modeling Neuron Selectivity \\ for Image Classification }

\icmlauthor{Shu Kong}{aimerykong@gmail.com}
\icmlauthor{Zhuolin Jiang}{zhuolin.jiang@huawei.com}
\icmlauthor{Qiang Yang}{qiang.yang@huawei.com}

\icmlkeywords{Deep Learning, Feature Learning, Image Classification, Neuron Selectivity}

\vskip 0.3in
]

\begin{abstract}

We now know that mid-level features can greatly enhance the performance of image learning,
but how to automatically learn the image features efficiently and in an unsupervised manner is still an open question.
In this paper,
we present a very efficient mid-level feature learning approach (MidFea),
which only involves simple operations such as $k$-means clustering, convolution, pooling, vector quantization and random projection.
We explain why this simple method generates the desired features,
and argue that there might be no need to spend much time in learning low-level feature extractors.
Furthermore,
to boost the performance,
we propose to model the neuron selectivity (NS) principle by building an additional layer over the mid-level features before feeding the features into the classifier.
We show that the NS-layer learns category-specific neurons with both bottom-up inference and top-down analysis,
and thus supports fast inference for a query image.
We run extensive experiments on several public databases to demonstrate that
our approach can achieve state-of-the-art performances for face recognition, gender classification, age estimation and object categorization.
In particular,
we demonstrate that our approach is more than an order of magnitude faster than some recently proposed sparse coding based methods.

\end{abstract}

\section{Introduction}
\label{sec:intro}

Image classification performance relies on the quality of image features.
The  low-level features include the common hand-crafted ones such as SIFT~\cite{lowe2004distinctive} and HOG~\cite{dalal2005histograms},
and the learned ones from the building blocks in an unsupervised model, such as  Convolutional Deep Belief Networks (CDBN)~\cite{lee2009convolutional} and  Deconvolutional Networks (DN)~\cite{zeiler2011adaptive}.
We then take these features to improve the classification performance by generating mid-level features from the low-level ones through further operations such as sparse coding and pooling~\cite{lazebnik2006beyond, jarrett2009best, boureau2010learning}.

{
\begin{figure*}[t]
\centering	
\includegraphics[width=0.99\textwidth]{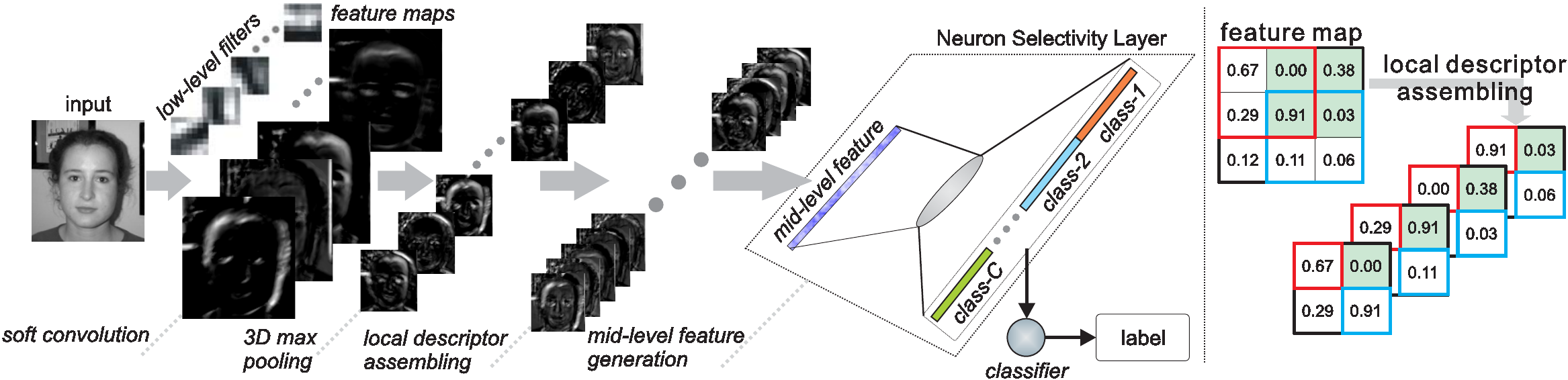} 
\caption{
Left panel: the overall flowchart of the proposed framework.
Specifically,
the proposed feed-forward \emph{MidFea} learns mid-level features in a hierarchical architecture,
then the Neuron Selectivity (\emph{NS}) layer transforms the features into a high-level semantic representation
which is fed into the linear classifier.
Right panel: demonstration of local descriptor assembling with $2\times2$ window (color image).
With the help of 3D max-pooling,
our local descriptor captures the salient orientations within a cuboid.
}
\label{fig:flowchart}
\end{figure*}
}

To learn the mid-level features,
these methods usually make use of a hierarchical architecture~\cite{zeiler2011adaptive},
in which each layer accumulates information from the layer beneath to form more complex features.
Despite their similarity,
they mainly differ in the design of nonlinearity,
which is the most important part for good classification performance~\cite{jarrett2009best}.
Spatial pyramid matching (SPM) based methods~\cite{yang2009linear, wang2010locality, boureau2010learning} apply sparse coding and max-pooling for the nonlinearity.
DN focuses on sparse coding, pooling and unpooling  for the nonlinearity~\cite{zeiler2011adaptive}.
CDBN uses sparse coding and quasi max-pooling~\cite{lee2009convolutional}.
Predictive Sparse Decomposition (PSD) further introduces nonlinear absolute value rectification and local contrast normalization~\cite{jarrett2009best}.
However,
as pointed out by Coates \etal~\cite{coates2011analysis},  
while some feature-learning approaches are slightly better than others,
it is not the difference of these methods that leads to an accuracy gain.
Moreover, complex methods can be easily outweighed by simpler ones that carefully consider some specific factors,
such as the receptive field size and density of extracted low-level features.
Therefore,
instead of designing complicated  approaches to learn mid-level features,
in this paper,
we propose an efficient mid-level feature learning method (\emph{MidFea}) which consists of very simple operations,
such as $k$-means, convolution, pooling, vector quantization and random projection as shown in Fig.~\ref{fig:flowchart}.
Through comparison with SIFT and HMAX~\cite{riesenhuber1999hierarchical},
we explain why our MidFea produces desirable features,
and argue that there might be no need to spend much time in learning low-level feature descriptors.

We also consider how to exploit the mid-level features more effectively to boost the learning performance.
According to studies in neural science~\cite{bienenstock1982theory},
neurons tend to selectively respond to visual signals from specific categories. 
Hence,
we build an additional Neuron-Selectivity (NS) layer over the mid-level features as demonstrated in Fig.~\ref{fig:flowchart},
so that the neurons can be fired selectively and semantically for the signals from specific categories.
By modeling this property as a structured sparse learning problem that supports both top-down analysis and bottom-up inference,
the NS-layer improves the  performance notably.

In summary, our contributions are two-fold.
(1)
We propose a simple and efficient method to learn mid-level features.
We give the explanation why our approach generates desirable features,
and argue that there might be no need to spend much time on learning the low-level features.
(2) To the best of our knowledge,
this is the first time that  neuron selectivity (NS) is modeled over the mid-level features to boost classification performance. Our model builds an NS-layer to support fast inference, which is an appealing property in real-world application.
We run extensive experiments to demonstrate our framework not only achieves state-of-the-art results on several databases,
but also runs faster than related methods by more than an order of magnitude.
We begin with describing our mid-level feature learning approach in Section~\ref{sec:MidFea},
followed by the proposed NS-layer in Section~\ref{sec:NSlayer}.
We then present the experimental validation in Section~\ref{sec:exp} and conclude in Section~\ref{sec:conclusion}.

\section{Mid-Level Feature Learning}
\label{sec:MidFea}

\subsection{Background}
\label{ssec:Background}

The concept of mid-level features was first introduced in~\cite{boureau2010learning},
meaning that features built over low-level ones remain close to image-level information without any need for high-level structured image description.
Typically,
the mid-level features~\cite{yang2009linear, wang2010locality, boureau2010learning} are learned via sparse coding techniques over low-level hand-crafted ones,
such as SIFT~\cite{lowe2004distinctive} and HOG~\cite{dalal2005histograms}.
However,
despite the promising performance in accuracy,
extracting the low-level descriptors requires significant amounts of domain knowledge and human labor.
The computation is also time-consuming and lacks flexibility.
As a result, researchers have been searching for alternative methods to learn the features for the system to be both efficient and effective.

Some impressive unsupervised feature learning methods have been developed such as CDBN~\cite{lee2009convolutional}, DN~\cite{zeiler2011adaptive} and autoencoders~\cite{hinton2006reducing}.
Despite their differences,
however,
empirical validation~\cite{jarrett2009best, coates2011analysis} confirms several rules of thumb.
First,
it is the nonlinearity in mid-level feature learning that leads to improved performance.
Second,
even complicated feature learning methods are better than others,
other factors may help simpler algorithms outperform the complex ones,
including more densely extracted local descriptors and suitable receptive field size.
Third,
despite the differences of these methods,
they consistently learn low-level features that resemble Gabor filters,
and even the simplest $k$-means can produce those similar extractors~\cite{coates2011analysis}.

Inspired by these studies,
we present a very efficient mid-level feature learning approach, called \emph{MidFea},
which consists of \emph{soft convolution},
\emph{3D max-pooling}, \emph{local descriptor assembling},
and \emph{mid-level feature generation} as shown in Fig.~\ref{fig:flowchart}.
Different from other mid-level feature learning methods that adopts SIFT and HOG,
ours learns  more adaptive low-level and mid-level features,
and  performs faster.

\subsection{The Proposed MidFea Algorithm}
\textbf{Soft Convolution.}
Our MidFea first runs $k$-means clustering in the training set to generate the low-level feature extractors.
Once the filters are derived,
however,
we do not use them in complicated nonlinear functions~\cite{lee2009convolutional} or analytical sparse decompositions~\cite{jarrett2009best, zeiler2011adaptive}.
Instead,
we convolve the image with these filters to get the feature maps,
which can be seen as a 3rd-order tensor in left panel of Fig.~\ref{fig:flowchart}.
It is worth noting that,
different from the simple convolution,
ours is a \emph{soft convolution} (\emph{sConv}) that adaptively generates sparse feature maps.
sConv consists of several steps\footnote{
Concretely,
sConv first convolves the image with the low-level filters,
followed by normalization over  all the feature maps along the third mode into a comparative range;
then thresholds the maps element-wisely with their mean map and produces sparse ones;
finally normalizes the sparse maps along the third mode again for the sake of subsequent operations. We test several choices to threshold the maps, and find that using mean map to threshold all maps always produce good results.}: convolution, normalization and thresholding,
which are demonstrated by Fig.~\ref{fig:illumination_invariance} (more illustration can be found in appendix).

There are several advantages in sConv.
First,
its convolutional behavior just equals to exhaustively dealing with all possible patches in the image,
and amounts to a densest local descriptor extraction.
Second,
normalization along the third mode preserves local contrast information by counting statistic orientations,
thus makes the resultant maps more resistant to illumination changes, as shown in Fig.~\ref{fig:illumination_invariance}.
Third,
the sparse property means trivial information or background noises can be filtered out by thresholding.
This can be seen in Fig.~\ref{fig:displayCaltech101} through comparison with dense SIFT feature maps.

\textbf{3D Max-Pooling.}
We adopt the 3D max-pooling operation~\cite{zeiler2011adaptive}  to obtain further robustness over these sConv feature maps.
Suppose we have 9 filters that generate 9 feature maps,
then 3D max-pooling operates in  a cuboid with size $2\times2\times2$,
meaning non-overlapping $2\times2$ neighborhood  between every pair of previous feature maps at the same location.
The pooling leads to a single value that is the maximum within the volume.
In a macro perspective,
we get 36 new  maps, each of which has the half size compared with that of the previous ones.
This simple operation not only reduces the size of the feature maps,
but also further captures the most salient information in a 3D neighborhood by eliminating trivial orientations.
It is worth noting that DN~\cite{zeiler2011adaptive} also uses 3D max-pooling for nonlinearity.
However,
DN is a top-down analysis method which requires  much time to derive the feature maps by sparse coding.
Ours is a feed-forward one and thus performs very fast.

\textbf{Local Descriptor Assembling.}
Low-level local descriptor is now assembled over the resulted 36 sConv feature maps,
as demonstrated by the right panel of Fig.~\ref{fig:flowchart}.
In detail,
by splitting each feature map into overlapping $2\times2$ patches,
we can produce 4 times more maps\footnote{Hereafter,
we ignore the boundary effect for presentation convenience.}.
Hence,
for the 36 feature maps,
we can generate 144 new ones now.
To have a better perception of these feature maps,
please refer to the SIFT feature maps in sparse coding based SPM (ScSPM)~\cite{yang2009linear}.
If we densely extract SIFT descriptors for patches centered at every pixel,
then we generate 128 feature maps, each of which has the same size with the image.

\textbf{Mid-Level Feature Generation.}
We encode the descriptors by \emph{vector quantization} (VQ) over a dictionary learned before hand.
Then,
we use max-pooling on the VQ codes in predefined partitions of the image\footnote{For example,
the image for object categorization is partitioned in spatial-pyramid scales of $2^l\times2^l$ (for $l=0,1,2$)~\cite{yang2009linear}.
The partitions are different for different tasks,
details are presented in experiments.},
and concatenate pooled codes into a large vector as the image representation.
As the concatenated vector usually has more than ten thousands dimensions,
we use \emph{random projection}~\cite{vempala2004random} for dimensionality reduction,
and normalize the reduced vector to have unit length as the final mid-level feature.
Note that random project is cheap to perform, as it does not involve large matrix factorization opposed to other dimensionality reduction methods.
Moreover,
even random project does not improve discrimination of the reduced data,
it produce performance-guaranteed results~\cite{vempala2004random}.

\begin{figure}[t]
\footnotesize
\centering	
\includegraphics[width=0.47\textwidth]{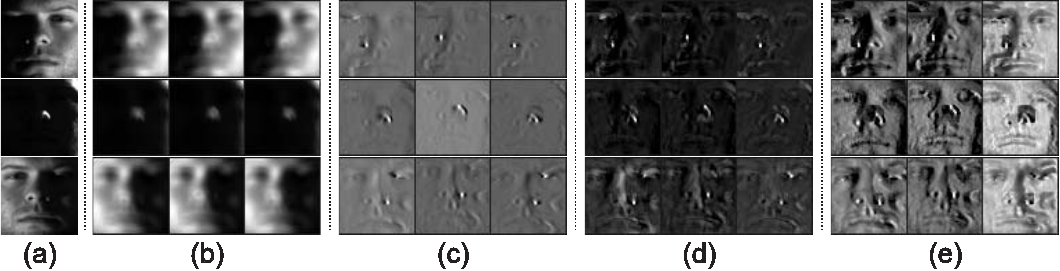}
\caption{Demonstration of soft convolution: (a) three images~\cite{EYaleB} of the same person under different illumination conditions;
(b) convolutional feature maps of each image displayed in each row with three different filters;
(c) normalized maps over (b);
(d) thresholded maps over (c);
and (e) normalized maps over (d).}
\label{fig:illumination_invariance}
\vspace{ -5mm}
\end{figure}

\subsection{Discussion}
In contrast to the hand-crafted  low-level descriptors such as SIFT and HMAX,
ours are learned adaptively within the data domain in an unsupervised manner.
Despite the main difference,
our model shares similarities with these hard-wired features.
For example,
SIFT captures eight fixed orientations over image patches,
while ours not only can do this,
but also captures subtle and important information due to its adaptivity and flexibility in the learning process.
Moreover,
our descriptor also resembles the HMAX feature,
which is derived in a feed-forward pathway and incorporate convolution and max-pooling.
But HMAX is built in a two-layer architecture and has no sparsity in the feature maps,
while ours produce more complicated and more resilient features in a deeper architecture with soft convolution.
Through the comparisons,
we can say these methods actually use the low-level descriptors to count the statistical information related to local orientations.
Therefore,
it may not need to incorporate low-level feature learning stage in a pipeline for high-level tasks,
let along learning them in a supervised manner.

Additionally, 3D max-pooling is also adopted in adaptive DN~\cite{zeiler2011adaptive} for nonlinear transformation.
However,
its sparse feature maps are calculated through convolutional sparse coding,
which means the maps have negative values that are hard to interpret.
Therefore,
adaptive DN only considers their absolute values.
On the contrary,
the proposed soft convolution provides non-negative elements for all feature maps,
so our model is more interpretable w.r.t capturing statistic information for the orientations.

\section{Neuron Selectivity Layer}
\label{sec:NSlayer}

Our mid-level features are generated in a purely unsupervised manner.
For the sake of classification,
we propose to build an additional layer over these mid-level features to boost the performance.
This layer models the neuron selectivity principle in neural science~\cite{bienenstock1982theory},
which means that certain neurons actively respond to the signals from a specific category, while others stay unfired.
Therefore, we call this layer Neuron Selectivity (NS) layer,
and its output  is fed into the classifier for classification.
Let $\x_i \in \RB^{p}$ denote the mid-level feature of the $i^{th}$ input image, which belongs to one of $C$ classes.
We would like to build an NS-layer with $d$ neurons.
Then,
the NS principle can be mathematically modeled as a structured sparse learning problem.

\subsection{Bottom-Up Inference and Top-Down Analysis}
Given a specific mid-level feature $\x_i$,
these NS-layer neurons selectively respond to the feature $\x_i$, and generate a set of activations  $\h_i \in \RB^{d}$.
We turn to an encoder function $f_{\W,{\bf b}}(\x_i)$,
where the filter $\W \in\RB^{d\times p} $ and  ${\bf b} \in\RB^{d}$,
to derive the activations $\h_i$.
In this paper,
we use the logistic function $\sigma (\cdot)$ to generate element-wise activations:
\begin{equation}\small
\begin{split}
\h_i = f_{\W,{\bf b}}(\x_i) = &\sigma (\W\x_i+\bf b).
\end{split}
\end{equation}

The encoder follows a bottom-up inference process,
and the produced activations are desired to have semantical structures.
In other words,
$\h_i$ should have specific sparse patterns for different class labels.
Before presenting how to produce the structured activations,
we first consider a top-down feedback analysis (decoder) from the activations with the inspiration in neural science~\cite{critchley2005neural} and successful applications in computer vision field~\cite{jiang2013label, zeiler2011adaptive}.
In this paper,
we choose the simple linear decoder to fulfill this goal:
\begin{equation} \small
\x_i \approx \D\h_i,
\end{equation}
where $\D\in\RB^{p\times d}$ is the weight matrix that controls the linear decoder.
Back to the idea of Neuron Selectivity which can be reflected by some appropriate constraints $\psi(\D)$ and $\phi(\H)$,
we unify the top-down analysis and bottom-up inference into one formulation as below:
\begin{equation}\small
\begin{split}
\min_{\D, \H, \W, {\bf b}} & \Vert \X - \D\H \Vert_F^2 +
\alpha \Vert \H - f_{\W, {\bf b}}(\X) \Vert_F^2, \\
& \text{s.t. }  \psi(\D), \phi(\H),
\end{split}
\end{equation}
where  $\X\in\RB^{p\times N}$ stacks all the $N$ training data in one matrix, $\H\in\RB^{d\times N }$ is the corresponding activations,
and $\alpha$ balances the effect of the two processes.

Note that the input mid-level features are normalized and  the encoder function is bounded in the range $[0, 1]$.
Therefore,
without losing generality,
by considering the decoder as a linear combination of  bases in $\D$ to reconstruct the mid-level features,
we constrain the columns in $\D$ to have unit Euclidean length,
\ie $\Vert \D_i\Vert_2^2 = 1, \forall i = 1, \dots, d.$

The  constraint on $\H$ is crucial and reflects the neuron selectivity property.
Hence,
we enforce a class-specific constraint on the activations.
In other words,
a particular set of neurons should actively respond to signals from a specific class,
while others stay unfired.
This property can be modeled as a structured sparse learning problem.
Instead of explicitly allocating the neurons to each class~\cite{yang2011fisher, kong2012dictionary, jiang2013label}, 
we implicit model this property via imposing an $\ell_{2,1}$ norm to eliminate the rows in $\H_c$,
\ie $\Vert \H_c \Vert_{2,1} = \sum_{j=1}^{d} \Vert\H_c^{(j)}\Vert_2$.

Besides,
the activations from the same class should resemble each other,
while those from different classes ought to be as different as possible.
To this end,
for the activations from the same class,
say the $c^{th}$ class denoted by $\H_c$,
we force them to be similar by minimizing
$\Vert \H_c - \bar\H_c\Vert_F^2$,
where $\bar\H_c$ is the mean vector matrix (by taking the mean vector of activations $\H_c$ as its columns) of activations from the $c^{th}$ class.
At the same time,
to differentiate the activations from different classes,
we drive the activations as independent as possible at class level by minimizing $\sum_{c=1}^{C}\Vert \H_c^T \H_{/c}\Vert_F^2$,
where $\H_{/c} = [\H_1, \dots, \H_{c-1}, \H_{c+1}, \dots, \H_{C}]$.
Taking  the constraints on $\H$ as a Lagrangian multiplier,
we have:
\begin{equation}\small
\phi(\H) =   \sum_{c=1}^{C}  \Big\{ \lambda \Vert \H_{c} \Vert_{2,1}
+ \beta\Vert \H_c - \bar\H_c\Vert_F^2 + \gamma\Vert \H_c^T \H_{/c}\Vert_F^2 \Big\},
\nonumber
\end{equation}
where  parameters $\lambda$, $\beta$ and $\gamma$ control each penalty term.

The constraint defined by $\phi(\H)$ has several meanings.
With the help of the first term,
a sufficiently large $\gamma$ decays the third term to zero.
It means the neurons automatically break into $C$ separate parts,
and each part corresponds to only one specific class.
When $\beta$ is large enough,
the neurons will respond to stimuli from the same class in an identical behavior.
This means the second term enforces the mechanism to be a strong classifier.
Instead of forcing the penalty so rigorously,
we set the three parameters in $\phi(\H)$ to proper values,
and allow:
(1) the intra-class variance to be preserved to prevent overfitting in training process,
and
(2) a few neurons to be shared across categories so that the combination of fired neurons ensures both discrimination and compactness.
Now,
we arrive at our final objective function with the Lagrangian multiplier $\phi(\H)$:
\begin{equation}\small
\begin{split}
\min_{\D, \H, \W} & \Vert \X - \D\H \Vert_F^2 +
\alpha \Vert \H - f_{\W,{\bf b}}(\X) \Vert_F^2
+ \phi(\H)\\
&\text{s.t. } \Vert \D_i\Vert_2^2 = 1, \forall i = 1, \dots, d.
\end{split}
\label{eq:obj}
\end{equation}
Each variable in Eq.~\ref{eq:obj} can be alternatively optimized through gradient descent method by fixing the others.
For detailed derivation and optimization,
please refer to appendix.

\subsection{Discussion}
Mathematically,
the proposed NS-layer,
reflected by the objective function in Eq.~\ref{eq:obj},
can be seen as a fast inference for sparse coding~\cite{jarrett2009best}.
Different from these methods,
ours predicts the structured sparse codes~\cite{topofilterCVPR2009}.
Moreover,
the decoder term with the constraints can be seen as discriminative dictionary learning~\cite{yang2011fisher, kong2012dictionary, jiang2013label, KongPR2014},
which will improve the classification performance in an analytical manner.
Because of the joint Sigmoid encoder,
the sparse codes are forced to be non-negative.
This is a desirable property that models the intuition of combining parts to form a whole,
as opposed to the classic sparse coding which includes negative elements to cancel each other out~\cite{hoyer2002non}.

\section{Experiments}
\label{sec:exp}
In this section,
we run extensive experiments to evaluate the proposed MidFea and NS-layer in feature learning and image classification performance\footnote{Code is available at Shu Kong's GitHub: \emph{https://github.com/aimerykong}  }.
First,
we study our MidFea and NS-layer on classification accuracy gains in a controlled way.
Then,
we highlight the efficiency of our framework according to the inference time for object categorization.
Moreover,
we carry out classification comparisons on four datasets for different tasks.
Finally,
we discuss some important parameters in our framework.

Specifically,
for classification comparisons,
we first use a subset of AR database~\cite{martinez1998ar} for face recognition and gender classification.
AR database consists of 50 male and 50 female subjects,
and each subject has 14 images (resized to $66\times48$) captured in two sessions with illumination and expression changes.
For face recognition,
the first 7 images  in Session 1 of each person are used for training and the rest for testing;
while for gender classification,
the first 25 male and the first 25 female individuals are used for training and the rest for testing.
we also test our framework on age estimation over the FG-NET database~\cite{geng2007automatic} with images (resized to $60\times60$)  spanning the age from 0 to 69.
Consistent with the literature,
we use leave-one-person-out setting for the evaluation.
Finally,
we evaluate our framework on object categorization over Caltech101~\cite{fei2007learning} and Caltech256~\cite{griffin2007caltech}.

Throughout the experiments,
we use the linear SVM toolbox~\cite{chang2011libsvm} as the classifier,
and choose the same settings to learn the low-level features,
\ie adaptively learning 9 filters for soft convolution and each one is with size $7\times7$.
But the partitions for spatial pooling are different for different tasks,
we demonstrate this along with the experiments.
Moreover,
we use the classification accuracy for face recognition, gender classification and object categorization,
and the Mean Absolute Error (MAE) for age estimation.

\subsection{Accuracy Gains by MidFea and NS-Layer}

To demonstrate the superiority of our  MidFea and the NS-layer,
we compare our model  in a controlled way with self-taught (ST) learning method~\cite{raina2007self},
which can be seen as a three-layer network with the sparse codes as the mid-level features.
Face recognition over AR database is used for the comparison,
and tens of thousands face images (with alignment and rescale) downloaded from the internet are used for unsupervised feature learning.
For ST,
we vary the number of dictionary bases from 200 to 1200 and the number\footnote{$0$ means no unlabeled data available.
In this case, a random matrix is used as the dictionary.} of unlabeled face images from $0$ up to $80,000$.
We record in Fig.~\ref{fig:selftaught_demo} the classification accuracies,
as well as that obtained by linear SVM on the raw image.
From the figure,
we can see, consistent with the literature,
more neurons (bases) lead to better performance,
and more unlabeled data learns more reliable dictionary for ST.
But when sufficient unlabeled data are available to learn the dictionary with a certain amount of bases, the accuracy will eventually saturate.

However,
when we add our NS-layer over the mid-level features produced by ST,
a notable gain is obtained.
Moreover,
when the features are generated by our MidFea (with 500 codewords for VQ and a single layer of $3\times 3$ partition for spatial pooling),
much better performance is achieved.
With no surprise,
once NS-layer is further built over our MidFea  (\emph{MidFea-NS}),
the best performance is recorded.
We can conclude that,
the proposed MidFea  learns more robust features to represent the image,
and our NS-layer further boosts the final classification performance.

{
\begin{figure}[t]
\centering	
\includegraphics[width=0.4\textwidth]{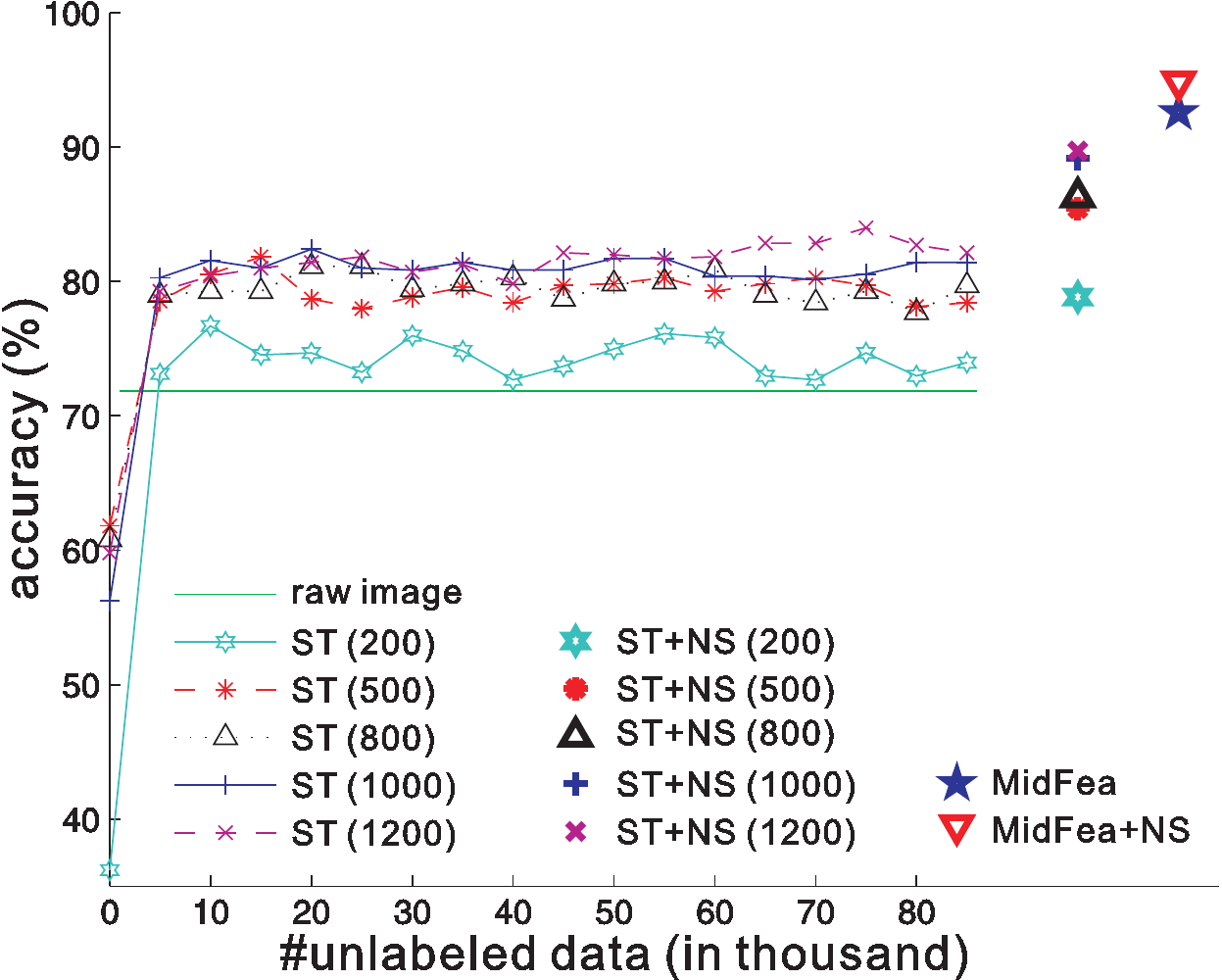}
\caption{ Demonstration of the proposed MidFea and the NS-layer in accuracy gains by comparison with self-taught learning (ST),
which can be seen as a three-layer network with the sparse codes as the mid-level features.
The $x$-axis indicates the amount of unlabeled data used for learning bases by ST.
The number in bracket shows the dimensions of the mid-level features in ST.
}
\label{fig:selftaught_demo}
\end{figure}
}

\subsection{Inference on Object Categorization}

\begin{table*}[t]
 \centering
  \caption{Comparison of detailed inference time (\emph{s}).
  VQ, SC and SP stand for vector quantization, sparse coding and spatial pyramid respectively.
For fair of comparison, the mid-level features of all the methods are reduced to 3000 dimension by random projection.   }
   \begin{tabular}{   l   l | c  | c  | c   c | c}
   \hline
        Time (\emph{ms})    & &   KSPM &  ScSPM & LLC  & &  \textbf{Ours} \\
    \hline\hline
        soft convolution       &&    \multicolumn{3}{c}{ \multirow{3}[0]{*}{ {\normalsize SIFT: $19.42 $  ($ 0.24$)} } }  && \textbf{0.070}  \\
        3D max-pooling       &&    \multicolumn{3}{c}{ \multirow{3}[0]{*}{   } }  & & \textbf{0.077}         \\
        local descriptor        &&    \multicolumn{3}{c}{ \multirow{3}[0]{*}{   } }  &   & \textbf{0.045}  \\
        \hline
        VQ or SC                   &&  0.46   &  $29.72$ &$47.06$    & & \textbf{0.174} \\
        \hline
        SP pooling                && \multicolumn{5}{c}{ { 0.114} } \\
        \hline
        random projection   && \multicolumn{5}{c}{ { 0.128} } \\
        \hline
        inference                  &&   0.75  &    \multicolumn{2}{c}{ { 0.63} }    & & \textbf{0.26}   \\
    \hline\hline
        \multicolumn{1}{c}{ \multirow{2}[0]{*}{ total time} } &&    20.87   & 50.01   &   67.35   & & \multicolumn{1}{c}{ \multirow{2}[0]{*}{ {\textbf{0.868} } } }  \\
                                                 &&     (1.69) & (30.83) &  (48.17)  & &  \\
    \hline
    \end{tabular}
  \label{tab:timings}
\end{table*}

To highlight the efficiency of our framework,
we study the inference time for a $150\times150$-pixel image on MATLAB in a PC with dual-core CPU,
2.50GHz, 32-bit OS and 2GB RAM.
The main steps in our framework include soft-threshold convolution,
3D pooling, local descriptor assembling, VQ,
spatial pyramid (SP) pooling,
random projection and the inference with the classifier.

As our MidFea learns features in a bottom-up manner,
it costs much less time than top-down methods such as adaptive DN.
Specifically,
adaptive DN needs more than 1 minute to produce all the feature maps and 2 more seconds with the VQ and kernel classifier.
This is much slower than ours by almost two orders of magnitude (see Table~\ref{tab:timings}),
as it involves multiple iterations for decomposing the image into multi-layer feature maps.
Additionally,
we compare our model with three feed-forward methods, Kernel SPM (KSPM)~\cite{lazebnik2006beyond},
ScSPM~\cite{yang2009linear} and LLC~\cite{wang2010locality}.
Table~\ref{tab:timings} summarizes the detailed comparisons,
which demonstrate our method performs much faster than the compared ones by more than one order of magnitude.
The three methods use original SIFT for the low-level feature descriptor,
and thus need more than 19 seconds to extract SIFT features in an image of $150\times150$ pixels.
Even using fast SIFT extraction approach~\cite{lazebnik2006beyond},
the local descriptor generation is still one time slower than ours.
Furthermore,
ScSPM and LLC adopt sparse coding and locality-restricted coding,
hence more running time is required.
Especially,
a sorting process is required before coding each local descriptor in LLC,
so it is much slower.
Considering the main steps of our model are amenable to parallelization and GPU-based implementation,
we expect its applications in the real world.

\subsection{Facial Attributes Recognition}

\begin{table*}[t]
    \centering
\caption{Accuracies ($\%$) of face recognition and gender classification on the AR database.}
    \begin{tabular}{l c c}
    \toprule
    Method      & Face Recognition  & Gender Classification \\
    \midrule
    SVM-raw     & $86.7$    & $91.3$    \\
    \midrule
    SRC ~\cite{wright2009robust}           & $90.3$    & $92.1$    \\
    LLC~\cite{wang2010locality}             & $91.6$    & $92.4$    \\
    FDDL~\cite{yang2011fisher}              & $92.0$    & $93.7$    \\
    LC-KSVD~\cite{jiang2013label}        & $91.9$    & $93.4$    \\
    \midrule
    \textbf{MidFea}                                                 & \textbf{93.3}    & $\textbf{96.1}$    \\
    \textbf{MidFea-NS}                                           & \textbf{94.7}    & $\textbf{98.3}$    \\
    \bottomrule
    \end{tabular}%
  \label{tab:AR}%
\end{table*}%

We now evaluate our model on facial attributes recognition:
face recognition and gender classification on AR database,
and age estimation on FG-NET database.
The linear SVM on the raw image acts as the baseline (\emph{SVM-raw}).

For fair comparison on AR database,
we choose several state-of-the-art methods as their source codes are  online available,
including SRC~\cite{wright2009robust},
FDDL~\cite{yang2011fisher},
LC-KSVD~\cite{jiang2013label}\footnote{The setup in~\cite{jiang2013label} is different from ours, as LC-KSVD originally choose 20 images per person for training and 6 for testing. We run the code in our work with the same setup as other methods.} and LLC~\cite{wang2010locality}.
We use the random face~\cite{wright2009robust} (300-dimension) as the input for the first three methods to reproduce the results.
Like our framework,
the output of LLC is also projected to 300-dimension with random matrix before fed into the linear SVM.
For face recognition and gender classification,
our MidFea learns mid-level features with 500 codewords for VQ and a single layer of $3\times 3$ partition for spatial pooling.
Moreover,
our NS-layer learns 300 and 10 neurons for the two tasks, respectively.
Detailed comparisons are listed in Table~\ref{tab:AR},
and some learned neurons w.r.t the two tasks are displayed in Fig.~\ref{fig:AR_filter}\footnote{To display the learned neurons,
hereafter we use the PCA for the dimensionality reduction other than random projection.
Moreover,
the spatial pooling is waived here,
the neurons are averaged w.r.t one image and then projected back to the input space for the sake of demonstration.
}.

From Table~\ref{tab:AR},
we can see with the proposed MidFea and NS-layer,
the performance outperforms the compared ones.
Furthermore,
as shown in Fig.~\ref{fig:AR_filter},
even the two tasks share the same database,
the learned neurons through the NS-layer capture  specific characteristics according to the task.
This intuitively demonstrates the reason why the proposed NS-layer works for classification.

{
\begin{figure}[t]
\centering	
    \begin{minipage}{0.450\textwidth}
		\centering
		\includegraphics[width=0.950\textwidth]{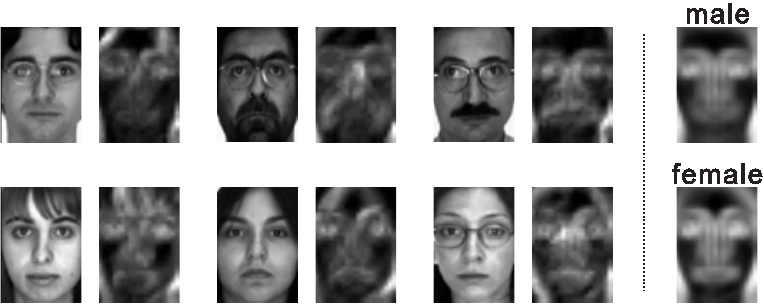}
	\end{minipage}
\caption{
Left panel: original images from AR dataset and the corresponding neurons at NS-layer for face recognition.
Right panel: the neurons  learned for gender classification.
}
\label{fig:AR_filter}
\end{figure}
}

{
\begin{figure}[t]
\centering	
		\includegraphics[width=0.450\textwidth]{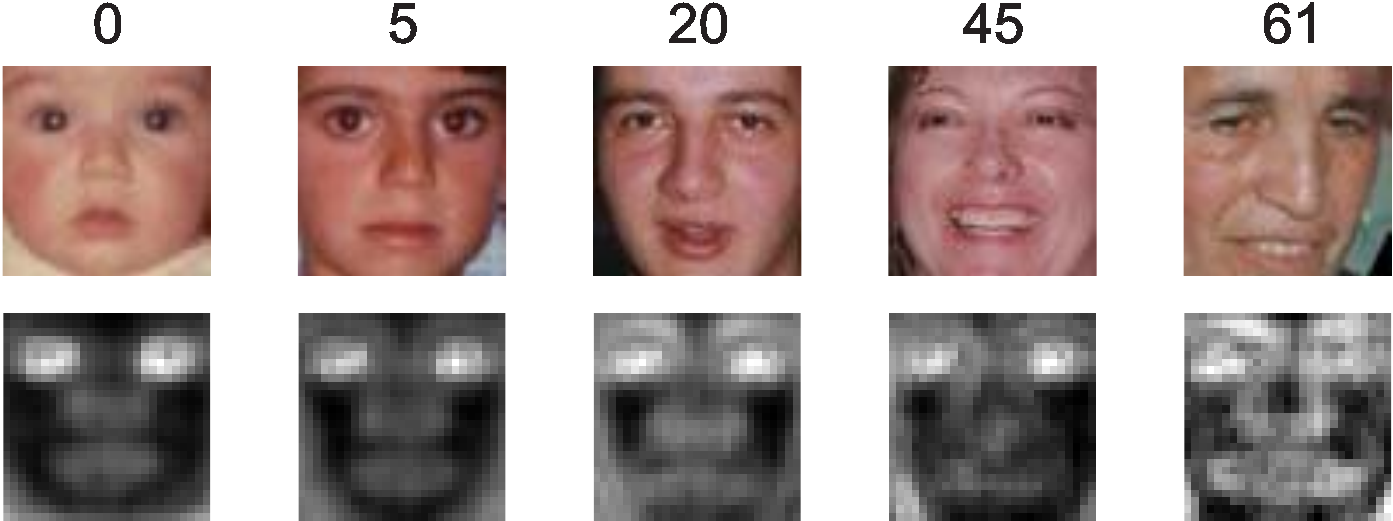}
\caption{
Neurons learned on the FG-NET for age estimation.
The response values reveal that different ages do have an association to specific neurons,
and we can see the neurons selectively response to the facial textures in older people as they have more wrinkles.
}
\label{fig:FGNET_filter}
\end{figure}
}

Additionally,
we use FG-NET database for age estimation.
Several state-of-the-art methods are compared here,
including
\emph{AGES}~\cite{geng2007automatic},
\emph{RUN}~\cite{yan2007learning},
\emph{OHRank}~\cite{chang2011ordinal},
\emph{MTWGP}~\cite{zhang2010},
\emph{BIF}~\cite{guo2009human},
and the recent \emph{CA-SVR}~\cite{chencumulative}.
Except for AGES,
all the methods uses the images with Active Appearance Model~\cite{cootes2001active}.
We generate 500-word codebook for VQ in MidFea,
and define the partition with a single layer of $8\times 8$-pixel overlapping grids for spatial pooling.
The results listed in Table~\ref{tab:ageEstimation} demonstrate our model performs slightly better than the best performance ever reported by the recent  CA-SVR,
which is sophisticatedly designed to deal with imbalanced data problem,
\eg there are very few images of  $60$ years old and above.
OHRank also deals with sparse data,
but performs very slow as showed in~\cite{chencumulative}.
The BIF method resembles ours as it uses the (hand-designed) biologically-inspired feature~\cite{riesenhuber1999hierarchical} for the mid-level features,
which, however, are generated in a shallower architecture.
Some neurons shown in Fig.~\ref{fig:FGNET_filter} demonstrate our model reveals the age information through the wrinkles on face.

\begin{table*}[t]
    \centering
\caption{MAE of age estimation on the FG-NET database.}
    \begin{tabular}{l c l c}
    \toprule
    Method      & MAE      &    Method      & MAE       \\
    \midrule
    SVM-raw     & $7.86$    & &\\
    \midrule
    AGES~\cite{geng2007automatic}        & $6.77$   &    RUN~\cite{yan2007learning}     & $5.78$    \\
    OHRank~\cite{chang2011ordinal}      & $4.85$    &   MTWGP~\cite{zhang2010}          & $4.83$    \\
    BIF~\cite{guo2009human}                  & $4.77$    &    CA-SVR~\cite{chencumulative}  & $4.67$    \\
    \midrule
    \textbf{MidFea}                                     & $\textbf{4.73}$    &    \textbf{MidFea-NS}                     & $\textbf{4.62}$    \\
    \bottomrule
    \end{tabular}
  \label{tab:ageEstimation}
\end{table*}

\subsection{Object Categorization}

{
\begin{figure*}[t]
\centering	
\includegraphics[width=0.40\textwidth]{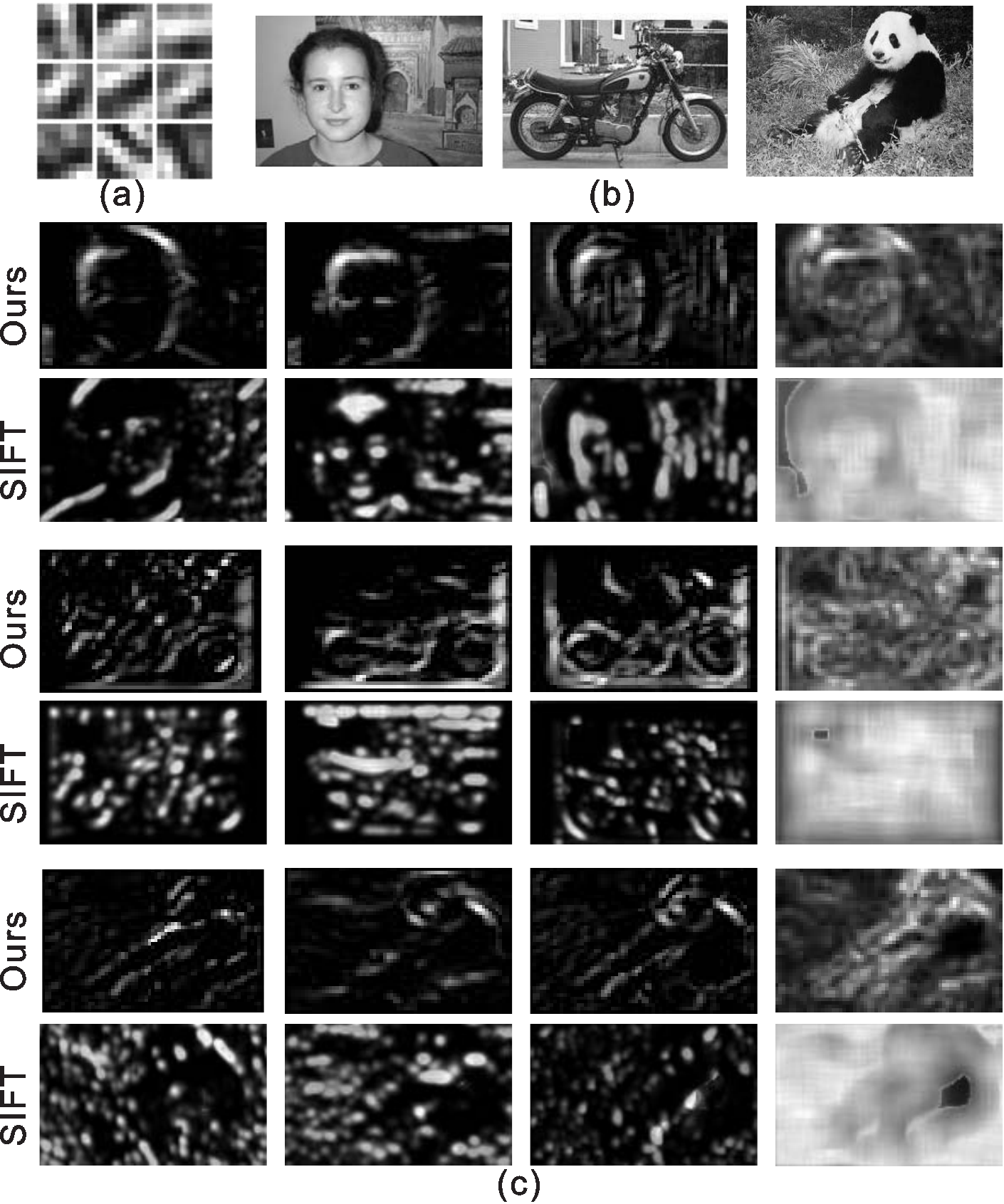}
\caption{ Visual comparison of local descriptor feature maps of our model and ScSPM~\cite{yang2009linear} for three images from Caltech101.
The nine learned filters are presented in (a).
(b) shows the three original images,
whose feature maps generated by our model (upper row) and ScSPM (bottom row) are presented in panel (c).
Note that the last image in each row is the average of all feature maps.
From the averaged map,
we can see the SIFT map distributes attention uniformly over the image,
while ours mainly focuses on the object.
}
\label{fig:displayCaltech101}
\end{figure*}
}

\begin{table*}[t]
    \centering
\caption{Accuracies and inference time over Caltech101.
For timing comparison, the mid-level features of all methods are reduced to 3000 dimension by random projection.
We observe this reduction does not effect the accuracies for these methods.
The time in bracket is achieved by  improved SIFT extraction method in~\cite{lazebnik2006beyond}.}
    \begin{tabular}{l r  l}
    \toprule
    Method      & ACC  ($\%$)  & Infer. Time (\emph{s})     \\
    \midrule
    Jarrett \etal (PSD)~\cite{jarrett2009best}                       & $65.6$    &   - \\
    Lee \etal (CDBN)~\cite{lee2009convolutional}              & $65.4$    &   - \\
    Zhang \etal~\cite{zhang2006svm}                                 & $66.2$    &   - \\
    Zeiler \etal (adaptive DN)~\cite{zeiler2011adaptive}    & $71.0$    &  $ 54.34$    \\
    \midrule
    Lazebnik \etal (SPM)~\cite{lazebnik2006beyond}         & $64.6$    &   $  21.25 $ ($2.08 $) \\
    Yang \etal (ScSPM)~\cite{yang2009linear}                    & $73.2$    &   $ 50.07 $ ($30.89$) \\
    Wang \etal (LLC)~\cite{wang2010locality}                     & $73.4$    &   $ 67.41 $ ($48.23$)\\
    Jiang \etal (LC-KSVD)~\cite{jiang2013label}                  & $73.6$    &   $ 25.90 $ ($6.72$)  \\
    \midrule
    \textbf{MidFea}                                                                 & $\textbf{73.8}$    &  \textbf{0.69}  \\
    \textbf{MidFea-NS}                                                           & $\textbf{74.7}$    &  \textbf{0.70}  \\
    \bottomrule
    \end{tabular}
  \label{tab:Caltech101}
\end{table*}

\begin{table}[t]
    \centering
\caption{Accuracies of object categorization on Caltech256.}
    \begin{tabular}{l l }
    \toprule
    Method      & ACC  ($\%$)     \\
    \midrule
    KSPM~\cite{yang2009linear}                                         & $29.5 \pm 0.5$    \\
    Yang \etal (ScSPM)~\cite{yang2009linear}                   & $34.0 \pm 0.6$    \\
    Wang \etal (LLC)~\cite{wang2010locality}                   & $30.4$    \\
    Zeiler \etal (adaptive DN)~\cite{zeiler2011adaptive}  & $33.2 \pm 0.8$    \\
    Boiman \etal (NN)~\cite{boiman2008defense}            & $38.0$ \\
    Jiang \etal (LC-KSVD)~\cite{jiang2013label}               & $34.3$ \\
    \midrule
    \textbf{MidFea}                                                              & $\textbf{36.6}  \pm \textbf{0.5} $    \\
    \textbf{MidFea-NS}                                                        & $\textbf{38.8}  \pm \textbf{0.4} $    \\
    \bottomrule
    \end{tabular}
  \label{tab:Caltech256}
\end{table}

We also evaluate our model on two popular databases, Caltech101 and Caltech256.
For both databases,
we randomly select  30 images of each category for training and the rest for testing,
and each image is resized to no larger than $150\times150$-pixel resolution with preserved aspect ratio.
The compared methods include both recent unsupervised feature learning methods and well-known methods with hand-crafted SIFT features.
The former methods include CDBN~\cite{lee2009convolutional}, adaptive DN~\cite{zeiler2011adaptive},
and PSD~\cite{jarrett2009best}.
The latter include the KSPM~\cite{lazebnik2006beyond},
ScSPM~\cite{yang2009linear} and LLC~\cite{wang2010locality}\footnote{LLC exploits a much larger dictionary (2048/4096 in Caltech101/256) for coding, we run the code with the same setup in our work.}.
For all the methods,
mid-level features are generated with a 1000-word codebook for VQ or sparse coding,
then reduced to 3000 dimension by random projection\footnote{We try reducing the original data to 5000/4000/3000 dimension, and the performance does not drop at all than that on the original data. But when we reduce them to 2000/1000, the accuracy drops by approximate 5$\%$ and $10\%$. That is why we choose 3000 in our work.}.
In our model,
the NS-layer learns 2040 and 5120 neurons in total for the two database respectively,
assuming an average of 20 neurons associate to one specific category.
The classic 3-layer-pyramid partition ($l=0,1,2$) for pooling is used.
Detailed comparisons are listed in Table~\ref{tab:Caltech101} for Caltech101 with the test time on the same image and Table~\ref{tab:Caltech256} for Caltech256 with the standard deviations over 10 trials.

It is easy to see that our method outperforms those with the sophisticated SIFT descriptor.
Most importantly,
the inference speed of our model is much faster than the compared ones  by  more than an order of magnitude.
Actually,
we can assemble the SIFT descriptor of every possible patch in one image as 128 feature maps in ScSPM.
Therefore,
we can compare the feature maps between ScSPM and ours to intuitively see the superiority of our model.
Fig.~\ref{fig:displayCaltech101} (a) displays the learned low-level feature extractors,
(c) shows some feature maps (full feature maps presented in appendix) of three images in (b).
Furthermore,
we average all the feature maps and show the averaged one in the last column of Fig.~\ref{fig:displayCaltech101} (c).
It is easy to see SIFT feature maps in ScSPM incorporates more cluttered background,
while ours focuses more on the object and discards the noisy region to a large extent.
We attribute this to the proposed soft convolution step.

Admittedly,
better performances are reported in literature on the datasets.
For example,
$75.7\%$ accuracy is achieved on Caltech101 in~\cite{boureau2010learning} with SIFT feature, sparse coding and intersection kernel;
Sohn \etal~\cite{sohn2011efficient} obtain $77.8\%$ on Caltech101 with much larger codebook (4096 atoms), more complicated sparse coding technique and the SIFT feature;
Varma \etal~\cite{varma2007learning} achieve state-of-the-art on Caltech256 as reported in~\cite{boiman2008defense} with multiple hand-crafted feature types and sophisticated techniques w.r.t the dataset.
However,
our model achieves impressive performance by simply learning the features in a purely unsupervised manner from low-level to mid-level.
It is far from the optimal mechanism for the task of classification.
Therefore,
based on our framework,
more sophisticated task-driven methods can be combined to address specific vision-based problems.

\subsection{Parameter Discussion}
Now we discuss the crucial parameters in our model (Eq.~\ref{eq:obj}),
including $\alpha$, $\beta$, $\gamma$ and $\lambda$.
Moreover,
the number of neurons in the NS-layer is also studied.
Fig.~\ref{fig:parameterDiscussion}  (a) presents the curve of accuracy vs. each parameter on AR database for face classification (with same setting of face recognition experiment).
It is easy to see the classification accuracy is not sensitive to these parameters,
and remains stable in a large range of them.
As well,
these hyper-parameters reveal that the terms in the objective function indeed brings performance gains.
Furthermore,
we show the accuracies vs. the neuron number in NS-layer on gender classification in Fig.~\ref{fig:parameterDiscussion}  (b), as the data is sufficient for this task (with same setting of gender classification experiment).
We can see the accuracy peaks with a small number of neurons,
say 6.
This demonstrates the effectiveness of NS-layer that serves for classification at a high level.



{
\begin{figure}[t]
\centering	
\includegraphics[width=0.450\textwidth]{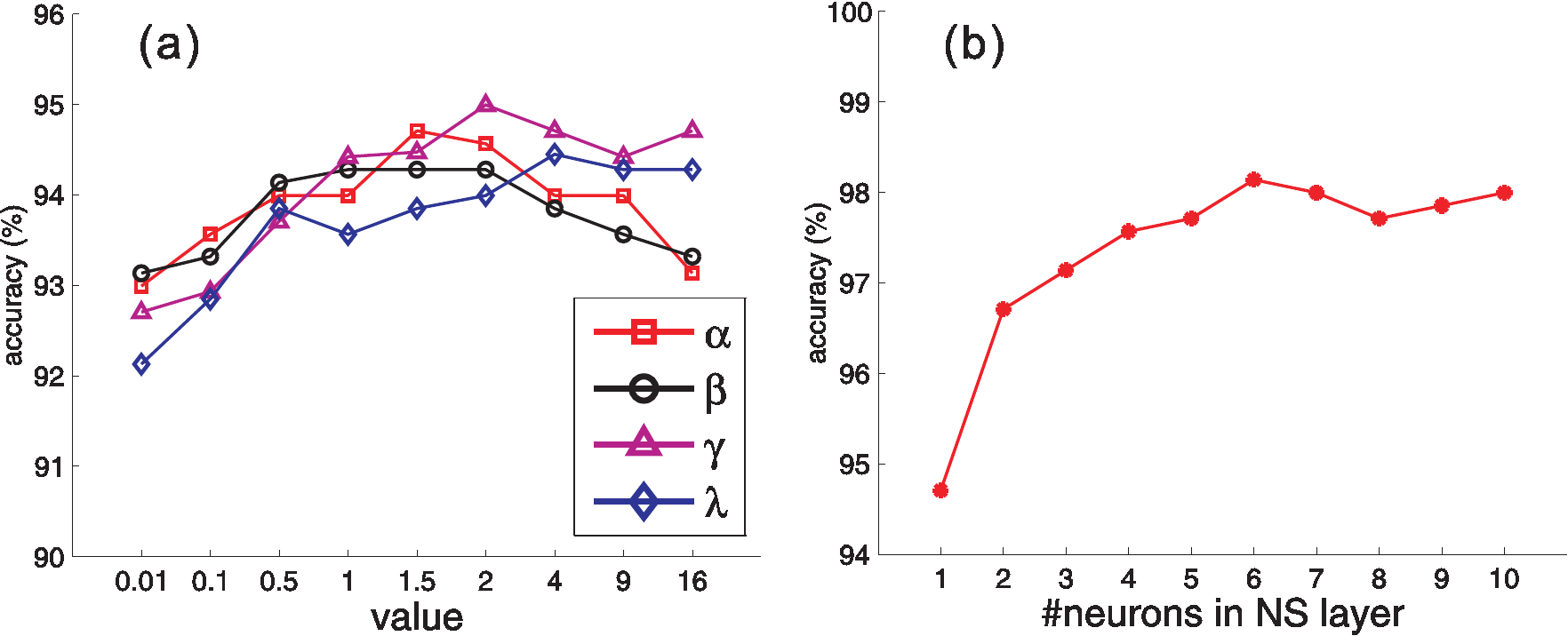}
\caption{
(a) The choices of $\alpha$, $\beta$, $\gamma$ and $\lambda$ vs. accuracy over AR database for face recognition;
(b) The number of neurons in NS-layer vs. accuracy over AR database for gender classification.
}
\label{fig:parameterDiscussion}
\end{figure}
}

\section{Conclusion with Future Work}
\label{sec:conclusion}

In this paper,
we present a simple and efficient method for learning mid-level features.
With comparison of the sophisticated hand-crafted features,
we explain why the proposed method produces the desired features.
We argue that there might be no need to spend much time in learning the low-level feature extractors.
Furthermore,
we propose to build an additional layer at higher level for classification.
The layer models the principle of neuron selectivity in neural science.
Given an image,
our MidFea produces the mid-level features very quickly in a feed-forward process,
and the NS-layer also supports fast bottom-up inference.
As a result,
our model performs faster than others by more than an order of magnitude,
and achieves comparable or even state-of-the-art classification performance as demonstrated by experiments.

Despite the effectiveness of the proposed feature learning method,
the performance gains in the public databases, especially the Caltech256 for object categorization,
still remains marginal.
The reason is obvious that the foreground objects are of large changes of appearance, translation and scale.
Therefore,
with more sophisticated features like~\cite{CNN4ImageNet, DeCAF} and our paper,
an explicit mechanism to deal with these changes is still solicited~\cite{KongCoRFL}.

{
\bibliography{example_paper}
\bibliographystyle{icml2014}
}

\section*{Appendix: Optimization at Neuron Selectivity Layer}

For presentational convenience,
we write the proposed objective function in Neuron Selectivity (NS) layer as below:
\begin{equation}\small
\begin{split}
\min_{\D, \H, \W,  {\bf b}} &  \Vert \X - \D\H \Vert_F^2 + \alpha \Vert \H - f_{\W,{\bf b}}(\X) \Vert_F^2 +  \\
\sum_{c=1}^{C} & \Big\{ \lambda \Vert \H_{c} \Vert_{2,1}
 + \beta\Vert \H_c - \bar\H_c\Vert_F^2 + \gamma\Vert \H_c^T \H_{/c}\Vert_F^2 \Big\}\\
& \text{s.t. } \Vert \D_i\Vert_2^2 = 1, \forall i = 1, \dots, d.
\end{split}
\label{eq:obj}
\end{equation}
Each variable in the above objective function can be alternatively optimized by fixing the others through stochastic gradient descent  method (SGD).

\subsection*{Updating $\D$}
Specifically,
we apply SGD  to update $\D$ by fixing the others,
and its gradient can be easily computed as below:
\begin{equation}\small
\nabla{\D} = -2\X\H^T + 2\D\H\H^T.
\label{eq:gradientD}
\end{equation}

Alternatively,
when bases number in $\D$ is not prohibitively large,
we can analytically update $\D=\X\H^T(\H\H^T)^{-1}$.
Then,
we normalize each column of $\D$ to have unit Euclidean length.
Note that this step will not pose any negative affects on the overall performance,
as the newly updated $\H$, on which $\D$ is only dependent,
can adaptively deal with this scaling change.

\subsection*{Updating $\H_c$ class by class}
Prior to optimizing $\H_c$,
which is the responses in the NS-layer to the data from the $c^{th}$ class,
we calculate the mean vector and get $\bar \H_c$ first.
Then, we update $\H_c$ as:
\begin{equation}\small
\begin{split}
\H_c^* = \argmin_{\H_c} & \Vert \X_c - \D\H_c \Vert_F^2 +
\alpha \Vert \H_c - f_{\W}(\X_c) \Vert_F^2\\
 + \beta\Vert  \bar\H_c &-  \H_c \Vert_F^2 + \gamma\Vert \H_{/c}^T\H_c \Vert_F^2  +\lambda_{\H} \Vert \H_{c} \Vert_{2,1},
\end{split}
\end{equation}
where $\X_c$ stacks all data from the $c^{th}$ class.
Let $\tilde\G_c = [ \X_c; \sqrt{\alpha}f_{\W}(\X_c); \sqrt{\beta}\bar\H_c; \0 ] \in \RB^{(p+2d+N-N_c)\times N_c}$,
and $\tilde\Q_c = [ \D; \sqrt{\alpha}\I ; \sqrt{\beta}\I; \sqrt{\gamma} \H_{/c}^T] \in \RB^{(p+2d+N-N_c)\times d}$,
in which $\0$ is a zero matrix with appropriate size.
We rewrite the above function as:
\begin{equation}\small
\begin{split}
g(\H_c) = \Vert \tilde\G_c - \tilde\Q_c\H_c \Vert_F^2 +\lambda_{\H} \Vert \H_{c} \Vert_{2,1}.
\end{split}
\label{eq:updateH}
\end{equation}
We use SGD to optimize $\H_c$,
and the partial derivative of $g$ w.r.t $\H_c$ is calculated as:
\begin{equation}\small
\begin{split}
\nabla \H_c = -2\tilde \Q_c^T \tilde\G_c + 2 \tilde\Q_c^T \tilde\Q_c\H_c + \lambda_{\H} \C\H_c,
\end{split}
\label{eq:gradientH}
\end{equation}
where $\C$ is a diagonal matrix with its $i^{th}$ diagonal element as:
\begin{equation}\small
\begin{split}
\C[i,i] = \frac{1}{\Vert\H_c^{(i)}\Vert_2},
\end{split}
\label{eq:diagC}
\end{equation}
where $\H_c^{(i)}$ is the $i^{th}$ row of $\H_c$.

Therefore, $\H_c$ can be optimized between solving Eq.~\ref{eq:diagC} and Eq.~\ref{eq:gradientH} for a couple of times.

\subsection*{Updating $\W$ and $\bf b$}

With the newly calculated $\H$,
we update $\W$ and $\bf b$ as:
\begin{equation}\small
\begin{split}
\{\W^*, {\bf b}^* \}= \argmin_{\W, {\bf b}} & \Vert \H - \sigma(\W\X + {\bf b}\1^T) \Vert_F^2,
\end{split}
\end{equation}
where $\1$ is a vector/matrix with all elements equaling 1 and appropriate size.
As $\W$ and $\bf b$ cannot be derived directly,
we use the SGD to update them.
With simple derivations,
by denoting the element-wise operation $\Si = \sigma(\W\X + {\bf b}\1^T)$,
we have the gradient of $\W$ and ${\bf b}$ as:
\begin{equation}
\begin{split}
\nabla{\W} = & 2 \big( (\Si - \H ) \odot \Si \odot (\1-\Si) \big) \X^T,\\
\nabla{\bf b} = &2 \big((\Si - \H )\odot\Si \odot(\1-\Si)\big) \1,
\end{split}
\label{eq:gradientWb}
\end{equation}
wherein "$\odot$" means Hadamard product.

\subsection*{Initialization and Algorithmic Summary }

Usually,
a good initialization for the variables can lead to fast convergence.
For example,
the linear decoder or the dictionary $\D$ can be pre-trained among each category.
Let $\D = [\D_1, \dots, \D_c, \dots, \D_C] \in \RB^{p\times d}$,
in which $\D_c$ are the neurons that only response to data from the $c^{th}$ class.
Then,
we can merely run $k$-means clustering among the data pool of class $c$ to obtain $\D_c$.
Then,
activations $\H_c$ can be initialized through the initialized $\D$.
Specifically,
for a datum $\x$,
we calculate a vector $\z \in \RB^{d}$ with its $i^{th}$ element as:
\begin{equation}
\begin{split}
z_i = \frac{ similarity({\bf d}_i, \x)}
            { \sum_{j=1}^{Ck} similarity({\bf d}_j, \x)},
\end{split}
\end{equation}
in which we can simply define $similarity(\m,\n) = \frac{1}{\Vert\m-\n\Vert_2}$.
After obtaining $\z$,
we get the initialized $\h = \frac{\z }{\Vert\z\Vert_2}$.
With the initialized $\H$,
both $\W$ and $\H$ can be then pre-trained for their initialization.

However,
the above initialization method lack flexibility,
because the allocation of $\D$ to each category must be pre-defined by hand.
Therefore,
we can also simply initialized all the variables with non-negative random matrices,
which serve the purpose of symmetry breaking.
Empirically,
we observe this random initialization does not mean inferior performance at all,
but requires more time to converge.
We owe it to that, even through our framework is a deep architecture,
the NS-layer only incorporates one hidden layer,
hence random initialization works quite well.

The overall steps (with gradient descent method)  of the NS-layer is summarized in Algorithm~\ref{alg:opt}.

 \begin{algorithm}[t]
\caption{ Algorithmic Summary at NS layer}
\begin{algorithmic}[1]
\REQUIRE training set,  $\X_c$ for class $c=1,\dots,C$
    \STATE initialize $\D$, $\H$, $\W$ and $\bf b$ randomly
    \WHILE{stop criterion is not reached}
        \STATE update $\D$ with its gradient in Eq.~\ref{eq:gradientD}
        \STATE update $\H_c$ with its gradient in Eq.~\ref{eq:gradientH}
        \STATE update $\W$ and $\bf b$ with their gradient in Eq.~\ref{eq:gradientWb}
    \ENDWHILE
\textbf{return} the learned $\W$, $\bf b$, $\D$
\end{algorithmic}
\label{alg:opt}
\end{algorithm}

\section*{Appendix: Highlight in Mid-Level Feature Learning}
Due to limited space of the paper,
we explain more advantages for the proposed mid-level feature learning method (\emph{MidFea}), especially the soft convolution (sConv).
As stated in the paper,
there are two significant advantages in sConv:
\begin{itemize}
  \item Normalization along the third-mode preserves local contrast information by counting statistic orientation, and thus makes the resultant feature maps more resistant to illumination changes (illustrated by Fig.~\ref{fig:EYaleB_ID1} to \ref{fig:EYaleB_ID3}).
  \item The sparse property means trivial information or background noises can be filtered out (illustrated by Fig.~\ref{fig:face} to \ref{fig:panda}).
\end{itemize}
Therefore,
in this section,
we illustrate the above two highlights through displaying full feature maps that appear in the paper.
All figures are preferably viewed by $300\%$ zooming in.

\subsection*{Illumination Invariance via Soft Convolution }
Illumination-invariant feature maps are obtained by the proposed soft convolution (sConv),
which only requires several simple operations: convolution, normalization and soft thresholding.
Put it in a pipeline,
sConv does the following:
\begin{enumerate}
  \item convoluting the original image with several low-level feature extractors (filters) and producing several (convolutional) feature maps, and stacks them in a third-order tensor;
  \item normalizing the feature maps along the third mode of the tensorial maps, scaling them into a comparative range for the sake of the subsequent soft thresholding operation;
  \item calculating the mean map along the third mode among the feature maps, and using it to threshold all the maps; (This stage produces sparse  feature maps.)
  \item employing a further normalization operation along the soft-threshold maps. (Similar to step 2, this stage will  benefit the subsequent local descriptor assembling operation.)
\end{enumerate}

\begin{figure}[t]
\footnotesize
\centering	
\includegraphics[width=0.1\textwidth]{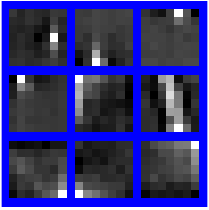}
\caption{ Low-level feature extractors (filters) for soft convolution.}
\label{fig:filter_EYaleB}
\end{figure}

To study this illumination invariance  property,
we use Extended YaleB database~\cite{georghiades2001few} for the demonstration.
This database contains 2,414 frontal face images of 38 individuals,
and is challenging due to varying illumination conditions and expressions.
Therefore, we use it to illustrate our proposed soft convolution for generating illumination-invariant feature maps.
Moreover,
the low-level feature extractors are shown in Fig.~\ref{fig:filter_EYaleB}.

\begin{figure*}[t]
\footnotesize
\centering	
\includegraphics[width=0.60\textwidth]{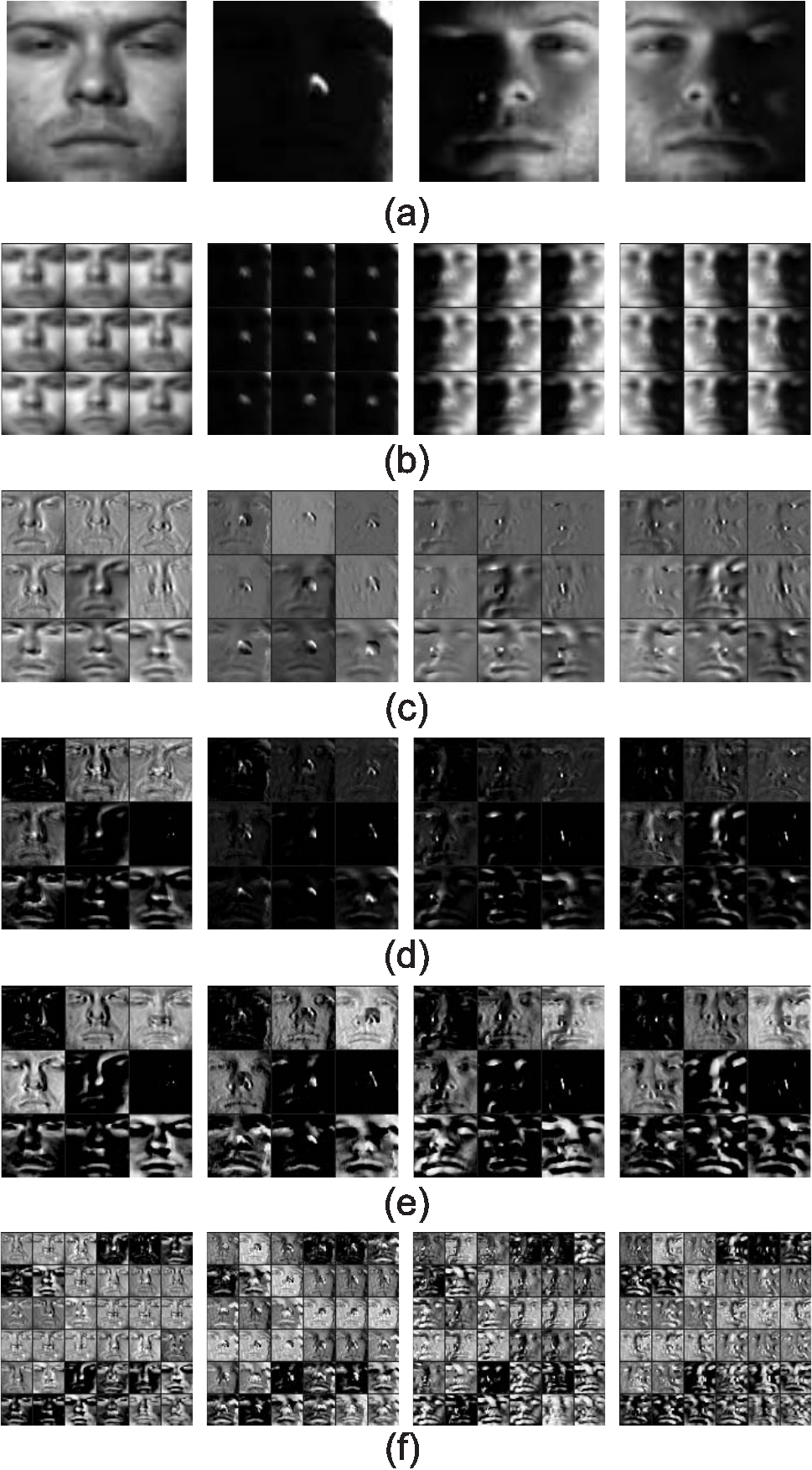}
\caption{ The first person. (a) original image, (b) convolutional feature maps, (c) normalization over convolutional feature maps, (d) soft-threshold maps, (e) normalization over the soft-threshold maps, (f) 3D max-pooling.}
\label{fig:EYaleB_ID1}
\end{figure*}

\begin{figure*}[t]
\footnotesize
\centering	
\includegraphics[width=0.60\textwidth]{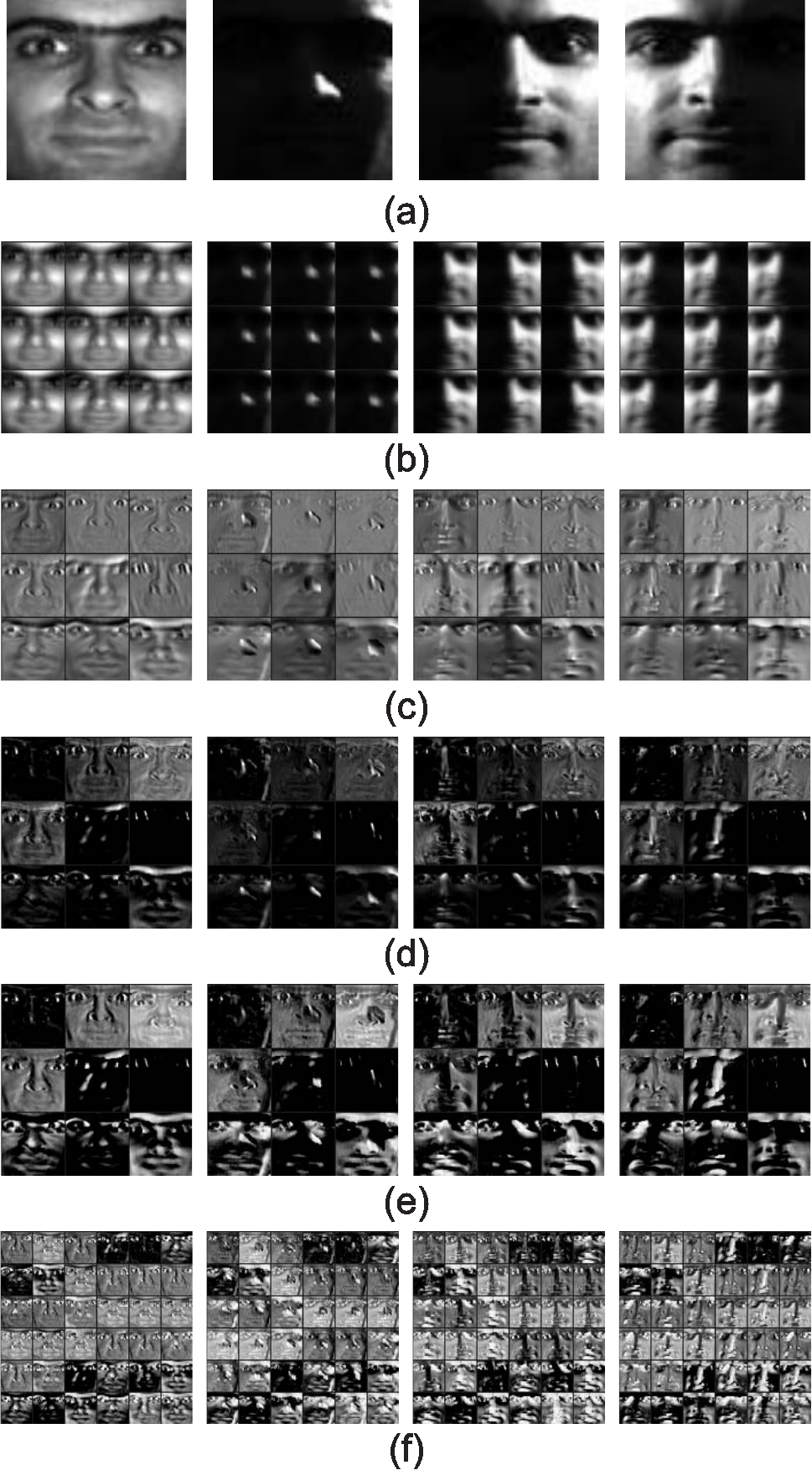}
\caption{ The second person.
(a) original image, (b) convolutional feature maps, (c) normalization over convolutional feature maps, (d) soft-threshold maps, (e) normalization over the soft-threshold maps, (f) 3D max-pooling.}
\label{fig:EYaleB_ID2}
\end{figure*}

\begin{figure*}[t]
\footnotesize
\centering	
\includegraphics[width=0.60\textwidth]{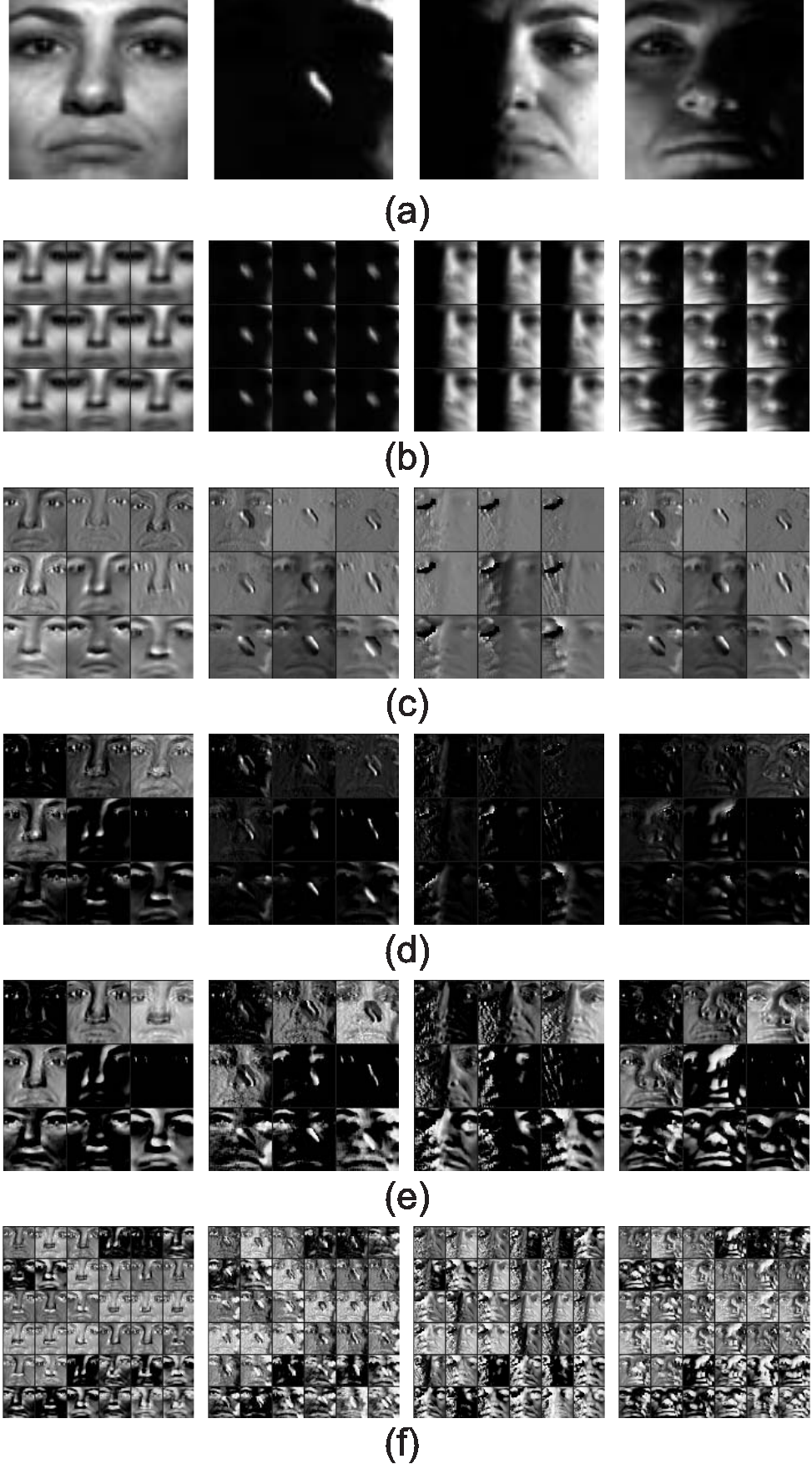}
\caption{ The third person.
(a) original image, (b) convolutional feature maps, (c) normalization over convolutional feature maps, (d) soft-threshold maps, (e) normalization over the soft-threshold maps, (f) 3D max-pooling.}
\label{fig:EYaleB_ID3}
\end{figure*}

We randomly select three different persons for demonstration,
and show three different images with different illumination for each person as Fig.~\ref{fig:EYaleB_ID1}, Fig.~\ref{fig:EYaleB_ID2} and Fig.~\ref{fig:EYaleB_ID3}.
In detail, in each figure, (a) to (f) are the original image,
the convolutional feature maps, normalized convolutional maps, soft-threshold maps, normalized maps over the soft-threshold ones, and 3D max-pooling feature maps, respectively.
It is easy to see the generated feature maps effectively handle the illumination changes.
In other word, the generated low-level feature (maps) is invariant to illumination.

We also evaluate our framework over this database for face recognition.
Consistent with the settings in the literature of face recognition on this database ,
half images per individual are randomly selected for training and the rest for test.
Our method leads to an average of $99.6\%$ accuracy.
To the best of our knowledge,
it is the best performance ever reported with this setting on the database.

\subsection*{Noise Filtering in Sparse Feature Maps }

The second highlight is that the generated feature maps can filter out background noises to some extent.
This property is brought out by the proposed soft convolution and 3D max-pooling.
In detail, trivial or weak orientation responses in the feature maps are removed through soft threshold operation,
while strong ones are kept and strengthened by the normalization.

\begin{figure*}[t]
\footnotesize
\centering	
\includegraphics[width=0.50\textwidth]{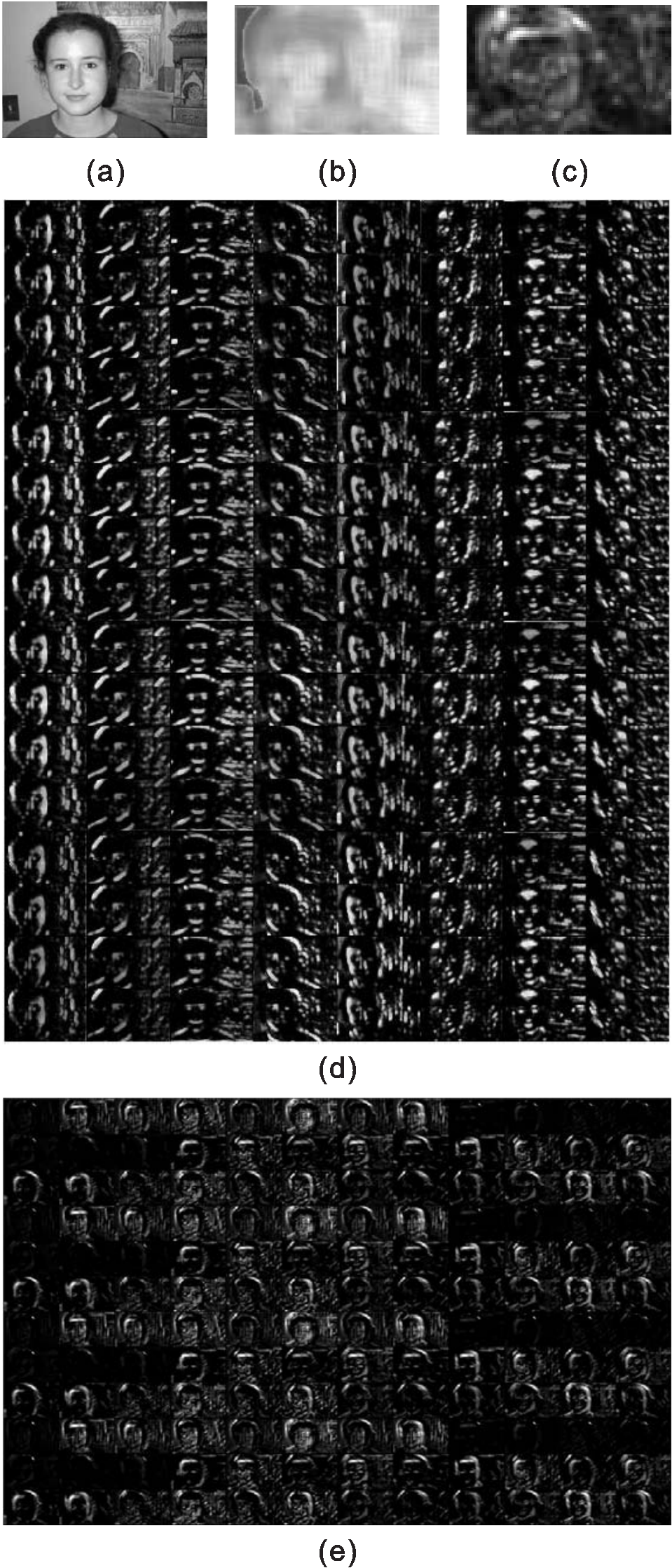}
\caption{ Faces image. (a) original image, (b) average over all maps of SIFT descriptor,
(c) average over all maps of the proposed local descriptor, (d) the full SIFT feature maps, (e) the full maps of the proposed descriptors. }
\label{fig:face}
\end{figure*}

\begin{figure*}[t]
\footnotesize
\centering	
\includegraphics[width=0.50\textwidth]{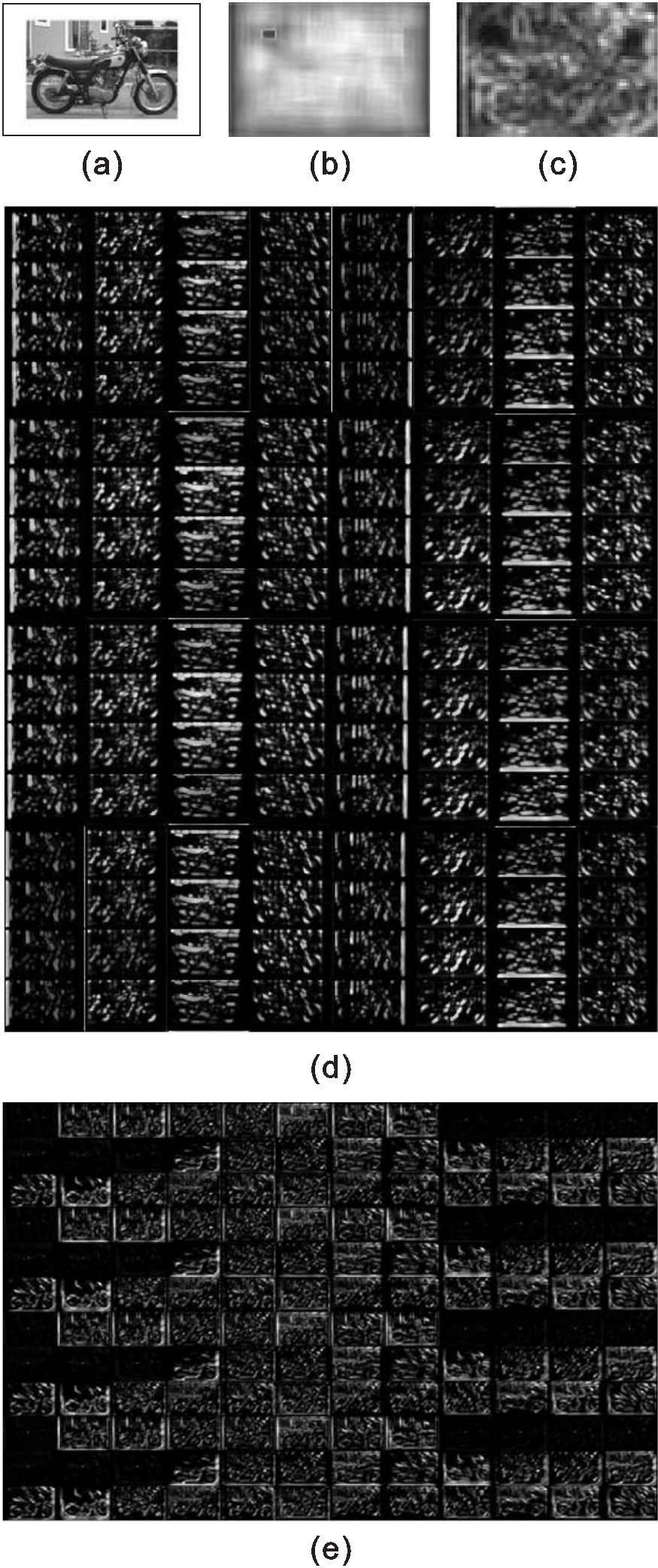}
\caption{ Motorbikes image. (a) original image, (b) average over all maps of SIFT descriptor,
(c) average over all maps of the proposed local descriptor, (d) the full SIFT feature maps, (e) the full maps of the proposed descriptors. }
\label{fig:moto}
\end{figure*}

\begin{figure*}[t]
\footnotesize
\centering	
\includegraphics[width=0.50\textwidth]{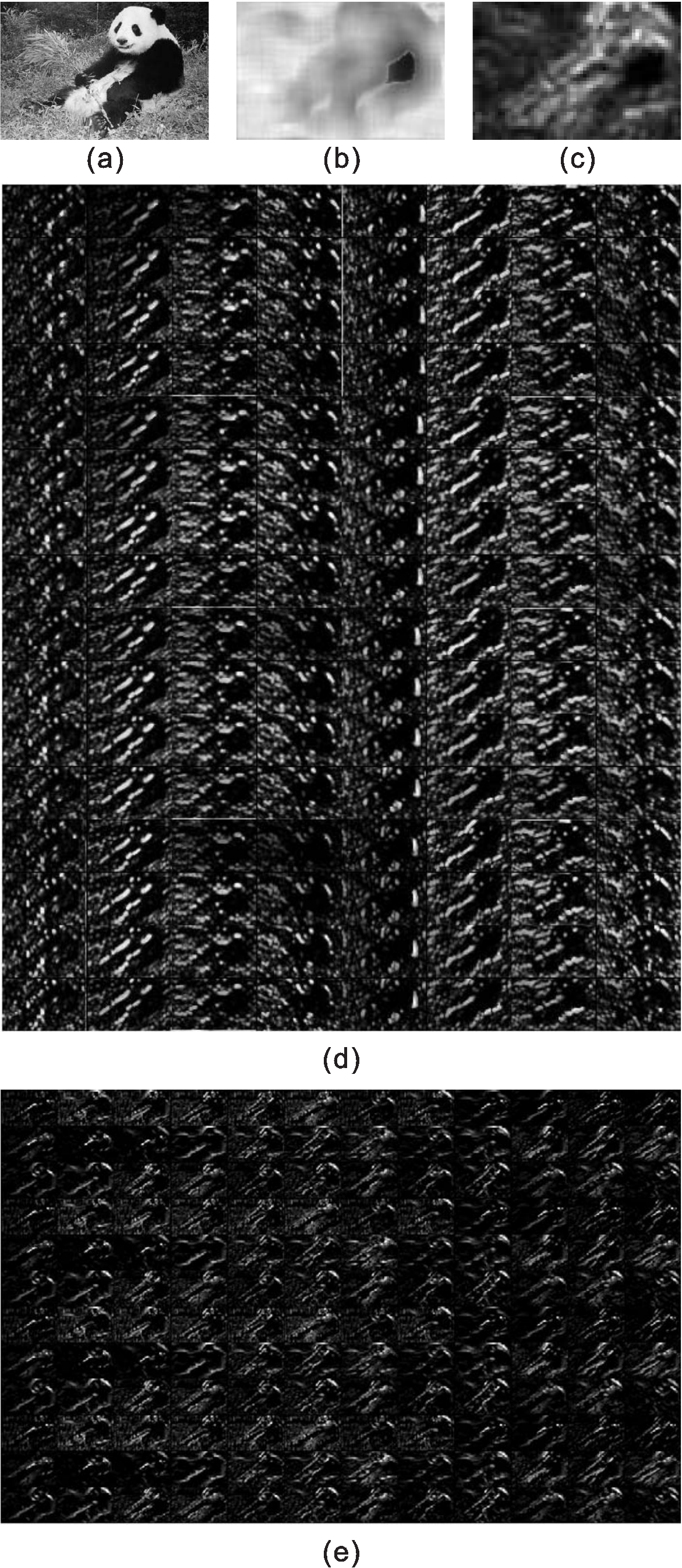}
\caption{ Panda image. (a) original image, (b) average over all maps of SIFT descriptor,
(c) average over all maps of the proposed local descriptor, (d) the full SIFT feature maps, (e) the full maps of the proposed descriptors. }
\label{fig:panda}
\end{figure*}

To show this property,
we display the full feature maps that appear in the experimental section of the paper.
Three images in Caltech101 database~\cite{fei2007learning} are randomly selected,
which come from three different categories,
Faces (image0117), Motorbikes (image0672) and panda (image0033).
Here,
the so-called feature maps consist of the assembled local descriptors (144 dimensions) of each patch.
Similarly,
the compared SIFT feature maps in SPM scheme consist of SIFT descriptors (128 dimensions) of all patches centered at every pixel in the image, therefore one image generates 128 feature maps.
The three sets of feature maps of the images are display in Fig.~\ref{fig:face}, Fig.~\ref{fig:moto} and Fig.~\ref{fig:panda}, respectively.
In each figure,
(a) shows the original gray-scale image,
(b) and (c) are the averaged feature maps of SIFT maps and ours that are demonstrated in (d) and (e),
respectively.
From the figures,
we can see the SIFT map distributes attention uniformly over the image,
while ours mainly focuses on the target object by constructing mid-level features.

\end{document}